\documentclass[10pt,twocolumn,letterpaper]{article}

\usepackage{cvpr}              

\usepackage{graphicx}
\usepackage{amsmath}
\usepackage{amssymb}
\usepackage{booktabs}

\usepackage{multirow}
\usepackage{makecell}

\usepackage[pagebackref,breaklinks,colorlinks]{hyperref}

\usepackage[capitalize]{cleveref}
\crefname{section}{Sec.}{Secs.}
\Crefname{section}{Section}{Sections}
\Crefname{table}{Table}{Tables}
\crefname{table}{Tab.}{Tabs.}

\def\cvprPaperID{5811} 
\def\confName{CVPR}
\def\confYear{2022}

\begin{document}

\title{High-Resolution Image Harmonization via Collaborative Dual Transformations}

\author{$\textnormal{Wenyan Cong}^{1}$, $\textnormal{Xinhao Tao}^{2}$, $\textnormal{Li Niu}^{1}$\thanks{Corresponding author.}, $\textnormal{Jing Liang}^{1}$, $\textnormal{Xuesong Gao}^{3,4}$, $\textnormal{Qihao Sun}^{4}$, $\textnormal{Liqing Zhang}^{1}$\\
$^1$ Shanghai Jiao Tong University 
$^2$ Harbin Institute of Technology
$^3$ Tianjin University
$^4$ Hisense\\
{\tt\small$^1$\{plcwyam17320,ustcnewly,leungjing\}@sjtu.edu.cn $^2$1180300213@stu.hit.edu.cn}\\
{\tt\small $^{3,4}$gaoxuesong@tju.edu.cn $^4$sunqihao@hisense.com $^1$zhang-lq@cs.sjtu.edu.cn }
}

\maketitle

\begin{abstract}
Given a composite image, image harmonization aims to adjust the foreground to make it compatible with the background. High-resolution image harmonization is in high demand, but still remains unexplored. Conventional image harmonization methods learn global RGB-to-RGB transformation which could effortlessly scale to high resolution, but ignore diverse local context. Recent deep learning methods learn the dense pixel-to-pixel transformation which could generate harmonious outputs, but are highly constrained in low resolution. In this work, we propose a high-resolution image harmonization network with Collaborative Dual Transformation (CDTNet) to combine pixel-to-pixel transformation and RGB-to-RGB transformation coherently in an end-to-end network. Our CDTNet consists of a low-resolution generator for pixel-to-pixel transformation, a color mapping module for RGB-to-RGB transformation, and a refinement module to take advantage of both. Extensive experiments on high-resolution benchmark dataset and our created high-resolution real composite images demonstrate that our CDTNet strikes a good balance between efficiency and effectiveness. Our used datasets can be found in \href{https://github.com/bcmi/CDTNet-High-Resolution-Image-Harmonization}{https://github.com/bcmi/CDTNet-High-Resolution-Image-Harmonization}.
\end{abstract}


\section{Introduction}\label{intro}

\begin{figure*}[t]
\centering
\subfloat[pixel-to-pixel transformation.]{
\label{fig:pix_trans}
\includegraphics[width=.297\linewidth]{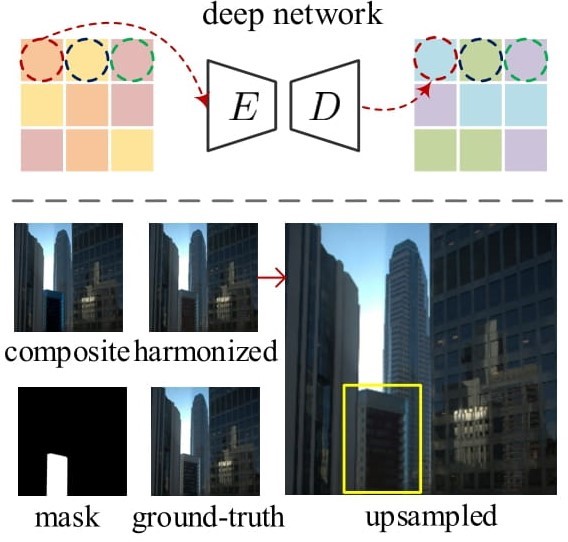}}
\quad
\subfloat[RGB-to-RGB transformation.]{
\label{fig:rgb_trans}
\includegraphics[width=.315\linewidth]{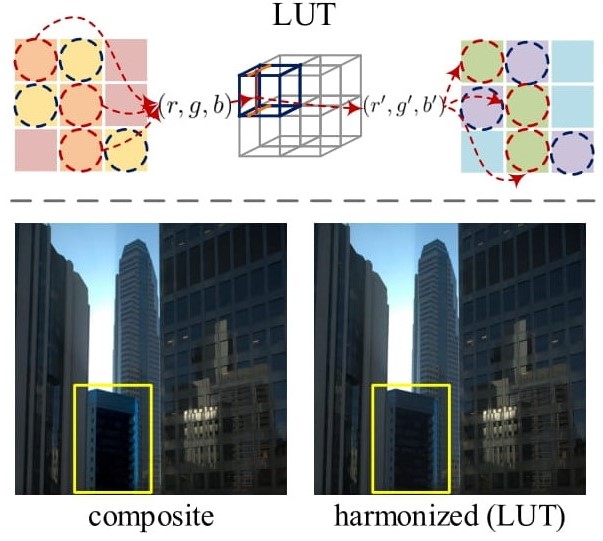}}
\quad
\subfloat[collaborative dual transformations.]{
\label{fig:our_trans}
\includegraphics[width=.31\linewidth]{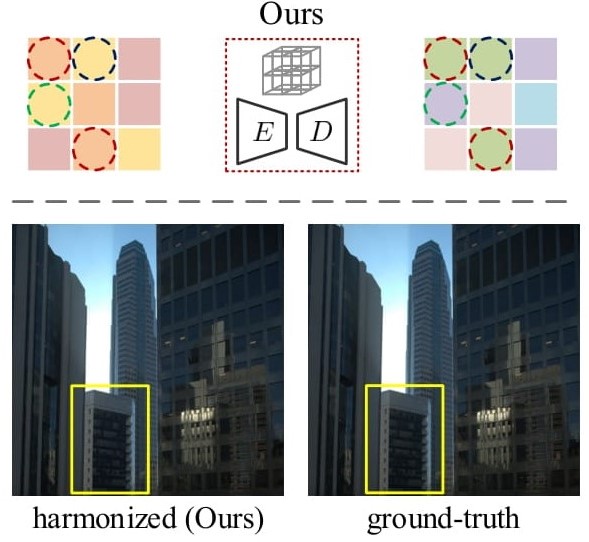}}
\caption{In \subref{fig:pix_trans}, deep harmonization network~\cite{sofiiuk2021foreground} learns dense transformation for each individual pixel and outputs low-resolution harmonious results, which would be blurry if upsampled. In \subref{fig:rgb_trans}, a 3D lookup table (LUT) in our method learns global color transformation and outputs high-resolution results without considering local context, which might lead to inharmonious local regions. In \subref{fig:our_trans}, our full method combines two transformations to achieve the most plausible results. Best viewed by zooming in.}
\label{fig:intro}
\end{figure*}

Image composition \cite{niu2021making} combines foreground and background from different images into a composite image. The quality of composite image may be degraded by the appearance (\emph{e.g.}, tone, illumination) inconsistency between foreground and background. To address this issue, image harmonization adjusts the foreground appearance to make it compatible with the background. Deep image harmonization methods \cite{tsai2017deep,DoveNet2020,bargain,sofiiuk2021foreground,guo2021intrinsic,ling2021region} have achieved remarkable progress by learning the dense pixel-to-pixel transformation between composite images and ground-truth harmonized images. However, they only performed low-resolution (\emph{e.g.}, $256\times256$) image harmonization, and a naïve upsampling can merely
lead to a large yet blurry output (see Figure \ref{fig:pix_trans}). 

Though directly training with high-resolution images could seemingly address this issue, the computational cost is very expensive.
For example, it will cost more than 950G FLOPs (floating point operations) and more than 20 GB memory for \cite{sofiiuk2021foreground} to harmonize a $2048\times 2048$ composite image. 
Besides, the high-resolution network may be weak in capturing long-range dependencies due to local convolution operations \cite{Wang_2018_CVPR} (see Section \ref{quanti} and the Supplementary). 


Prior to deep image harmonization, conventional harmonization methods \cite{reinhard2001color,multi-scale,xue2012understanding,lalonde2007using,zhu2015learning} mainly used hand-crafted statistical features (\emph{e.g.}, illumination, color temperature, contrast, saturation) to determine color-to-color transformation for foreground adjustment. Color-to-color transformation could be achieved in different color spaces, and we narrow the scope to RGB-to-RGB transformation in this work. RGB-to-RGB transformation is a global transformation of color values. Therefore, it is barely constrained by the number of pixels and could effortlessly scale to high resolution. However, global transformation  disregards the local context for each pixel, prone to generate harmonization results with local inharmony. 




Our key insight for high-resolution image harmonization is to combine the advantages of both pixel-to-pixel transformation and RGB-to-RGB transformation.  
We name our high-resolution image harmonization network with Collaborative Dual Transformations (CDT) as CDTNet.
CDTNet consists of a low-resolution generator for pixel-to-pixel transformation, a color mapping module for RGB-to-RGB transformation, and a refinement module to combine the best of two worlds. The low-resolution generator is a U-Net \cite{ronneberger2015u} structure which takes in a low-resolution composite image and outputs a low-resolution harmonized result. Meanwhile, the encoder feature is used to learn RGB-to-RGB transformation, unlike previous hand-crafted methods \cite{reinhard2001color,multi-scale}. Specifically, we learn several basis lookup tables (LUTs) \cite{LUT-inverse-halftoning3,LUT-image-enhancement1,LUT-image-denoising1,LUT-image-denoising2} shared by all images and a weight predictor based on the encoder feature to predict image-specific combination coefficients. RGB-to-RGB transformation is performed on high-resolution composite images using the combined LUT. With the RGB-to-RGB result as guidance, the refinement module utilizes both the harmonized result and decoder feature map from the low-resolution generator to compensate for fine-grained local information.

Considering the efficiency of CDTNet, pixel-to-pixel transformation only operates on low-resolution inputs and the refinement module is light-weighted, so the memory cost and computational cost are well-suppressed. Considering the effectiveness of CDTNet, RGB-to-RGB transformation could provide a holistic understanding of the whole image and the sharp edges of the transformed image, while pixel-to-pixel transformation could provide fine-grained local information. As shown in Figure \ref{fig:rgb_trans}, RGB-to-RGB transformation can obtain a globally reasonable tone and illumination. However, it may produce inharmonious local regions (\emph{i.e.}, the left and right borders of the foreground building) without considering local context. In contrast, our CDTNet can yield visually plausible and harmonious results of high resolution (see Figure \ref{fig:our_trans}). 
Our contributions could be summarized as follows: 

\begin{itemize}
\item  To the best of our knowledge, this is the first work focusing on high-resolution image harmonization.
\item We are the first to achieve deep learning based color-to-color transformation for image harmonization.
\item We unify pixel-to-pixel transformation and color-to-color transformation coherently in an end-to-end network named CDTNet.
\item Extensive experiments demonstrate that our CDTNet achieves state-of-the-art results with less resource consumption.
\end{itemize}

\section{Related Works}

\subsection{Image Harmonization}\label{rw-ih}
Traditional image harmonization methods ~\cite{colorharmonization,lalonde2007using,xue2012understanding,multi-scale,Pitie2005ndimensional,reinhard2001color,poisson,dragdroppaste,error-tolerant,zhu2015learning} mainly leveraged color-to-color transformation to match the visual appearance, which could be further divided into non-linear transformations \cite{multi-scale,xue2012understanding} and linear transformations \cite{reinhard2001color,lalonde2007using,zhu2015learning}.~\cite{multi-scale,xue2012understanding} proposed to match the pyramid histograms or histogram zones to address the appearance inconsistency.~\cite{reinhard2001color,lalonde2007using,zhu2015learning} applied a simple color adjustment to modify the foreground distribution by shifting and scaling the color values.

Recently, deep image harmonization methods \cite{DoveNet2020,xiaodong2019improving,ling2021region,Jiang_2021_ICCV,Guo_2021_ICCV} focused on learning dense pixel-to-pixel transformation from deep learning networks. \cite{tsai2017deep,sofiiuk2021foreground} both leveraged auxiliary semantic features to improve the basic image harmonization network. \cite{DoveNet2020} introduced a domain verification discriminator pulling close the foreground domain and background domain. \cite{xiaodong2019improving,Hao2020bmcv} explored various attention mechanisms for image harmonization. \cite{bargain,ling2021region} explicitly used background domain/style to guide the foreground harmonization. \cite{guo2021intrinsic} harmonized composite images by harmonizing reflectance and illumination separately. Different from existing methods, we focus on high-resolution image harmonization. Besides, instead of using only one type of transformation, we combine the complementary RGB-to-RGB transformation and pixel-to-pixel transformation into an end-to-end architecture.
\subsection{High-Resolution Image-to-Image Translation}\label{rw-i2it}

To the best of our knowledge, there are no previous works focusing on high-resolution image harmonization, but high-resolution image-to-image translation has been studied in many other fields like image segmentation \cite{LinRefineNet,chen2019GLNET}, image inpainting \cite{yi2020contextual}, image matting \cite{hdmatt}, style transfer \cite{lin2021drafting,anokhin2020high}, and image synthesis \cite{Chen2017,Wang2018,Liu_Zhu_Song_Elgammal_2021}.
Recent works could be mainly divided into three groups. The first group is placing a low-resolution generator embedded in a high-resolution generator. To name a few, \cite{Wang2018} pioneered the embedded scheme and extended the pix2pix to pix2pixHD to adapt to high-resolution applications. A slew of works followed this line and proposed Progressive GAN \cite{Karras2018} and its variants \cite{Andreini2019,Hamada2019}. \cite{Chen2017} employed a cascade of refinement modules to scale to high resolution. The second group is to stitch/merge low-resolution outputs. \cite{hdmatt} cropped the high-resolution image into patches and processed each patch with cross-patch consistency. \cite{anokhin2020high} shifted and downsampled the high-resolution image into multiple low-resolution images for separate processing. The third group leveraged deep learning techniques to predict color transformation \cite{gharbi2017deep,Zeng2020}, which is not constrained by the image resolution. 
Inspired by the third group of methods, this work is a pioneer in applying deep color-to-color transformation to image harmonization. Besides, collaborating with a low-resolution deep image harmonization network (\emph{i.e.}, pixel-to-pixel transformation) and a refinement module, our network can produce better high-resolution harmonization results with limited resources.

\section{Our Method}\label{method}

\begin{figure*}[tp!]
\begin{center}
\includegraphics[width=0.85\linewidth]{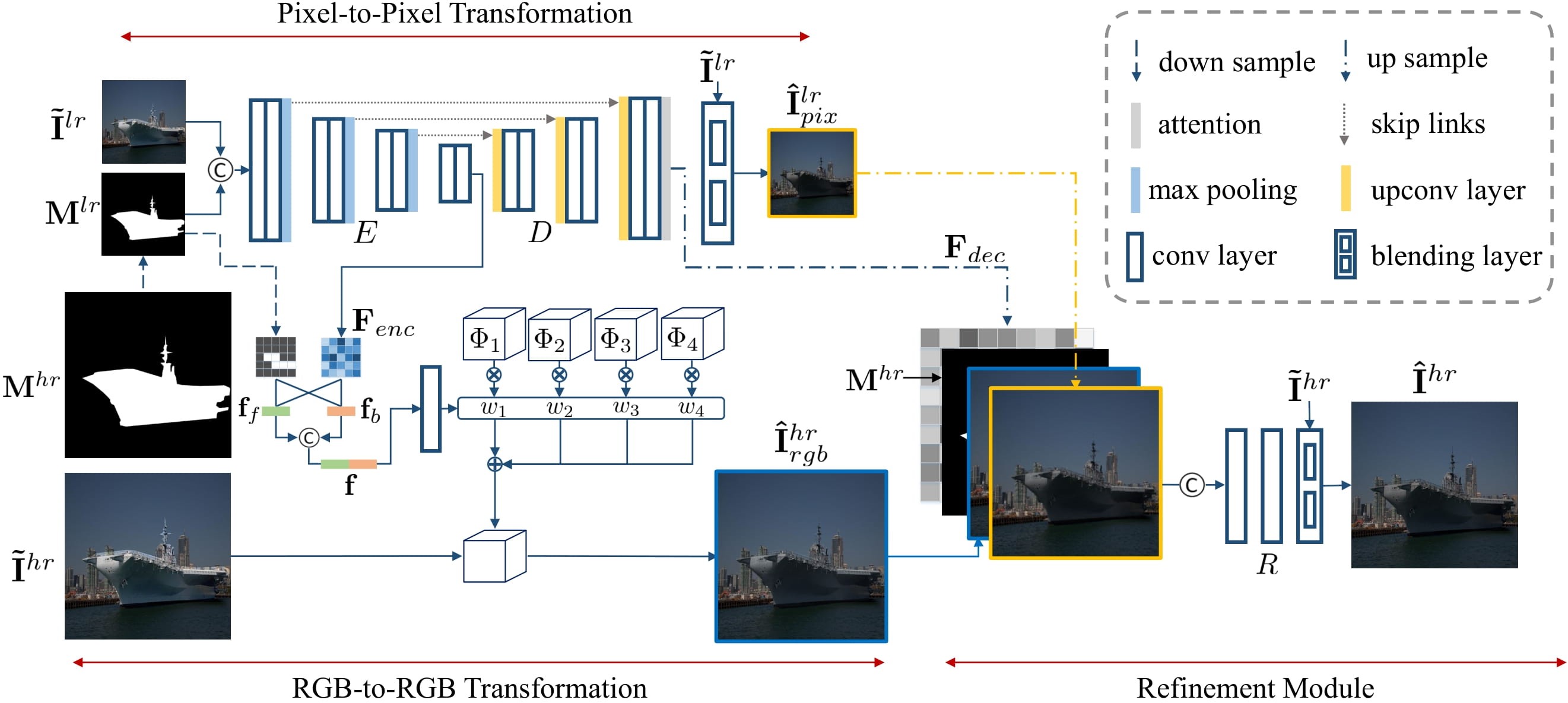}
\end{center}
\caption{The illustration of our CDTNet for high-resolution image harmonization, which takes in $(\tilde{\mathbf{I}}^{hr}, \mathbf{M}^{hr})$ and outputs $\hat{\mathbf{I}}^{hr}$. CDTNet contains a low-resolution generator (encoder $E$ and decoder $D$) for pixel-to-pixel transformation, a color mapping module (basis LUTs and weight predictor) for RGB-to-RGB transformation, and a refinement module $R$.}
\label{fig:network}
\end{figure*}

We propose a novel network, CDTNet, to reduce the computational burden and simultaneously maintain the harmonization performance. The pipeline of our CDTNet is shown in Figure \ref{fig:network}. Given a high-resolution composite image $\tilde{\mathbf{I}}^{hr}\in \mathbb{R}^{H\times W\times 3}$ and foreground mask $\mathbf{M}^{hr}$, image harmonization aims to obtain the harmonized result $\hat{\mathbf{I}}^{hr} \in \mathbb{R}^{H\times W\times 3}$.
We first downsample ($\tilde{\mathbf{I}}^{hr}$, $\mathbf{M}^{hr}$) to $h\times w$ (\emph{e.g.}, $h=w=256$) to obtain low-resolution $(\tilde{\mathbf{I}}^{lr}, \mathbf{M}^{lr})$. Our network contains three parts: a low-resolution generator, a color mapping module, and a light-weighted refinement module. The low-resolution image harmonization network is a U-Net-like architecture with encoder $E$ and decoder $D$, which takes in the low-resolution $(\tilde{\mathbf{I}}^{lr}, \mathbf{M}^{lr})$ and outputs low-resolution result $\hat{\mathbf{I}}^{lr}_{pix}$. The RGB-to-RGB transformation is built upon several basis transformations (\emph{i.e.}, LUTs) $\{\Phi_n\}_{n=1,...,N}$ and a weight predictor. The weight predictor takes the bottleneck feature map $\mathbf{F}_{enc}$ extracted from $E$ as input to predict the combination coefficients of basis transformations. After applying combined transformation, we could get a high-resolution output $\hat{\mathbf{I}}^{hr}_{rgb}$. Then, the upsampled $\hat{\mathbf{I}}^{lr}_{pix}$, the high-resolution output $\hat{\mathbf{I}}^{hr}_{rgb}$, the last decoder feature map $\mathbf{F}_{dec}$ from low-resolution generator, together with foreground mask $\mathbf{M}^{hr}$ are passed through refinement module $R$ to obtain a better harmonization output $\hat{\mathbf{I}}^{hr}$, which is expected to be close to the high-resolution ground-truth image $\mathbf{I}^{hr}$.

\subsection{Pixel-to-Pixel Transformation}\label{pix}

Dense pixel-to-pixel transformation has been widely explored by recent deep image harmonization methods. Pixel-to-pixel transformation is adept at adjusting each foreground pixel according to its local context, which could hardly be achieved by RGB-to-RGB transformation. To reduce the computational complexity and memory burden, we propose to leverage the deep image harmonization network with downsampled resolution.

The low-resolution harmonization network could be realized by any generator with an encoder-decoder structure. Recently in \cite{sofiiuk2021foreground}, they propose to apply an image blending layer to the last decoder feature map $\mathbf{F}_{dec}$ to obtain the harmonized foreground and the soft attention mask. Then the final output of the encoder-decoder network is obtained by blending harmonized foreground and input composite background using the soft attention mask.
They also propose a blending layer equipped generator named iS$^2$AM. Considering the simple structure and competitive performance of iS$^2$AM, we adopt it as the low-resolution generator responsible for our pixel-to-pixel transformation. As shown in Figure \ref{fig:network}, the low-resolution composite image and mask $(\tilde{\mathbf{I}}^{lr}, \mathbf{M}^{lr})$ are concatenated channel-wisely. After passing the input through the generator (with encoder $E$ and decoder $D$), we enforce the harmonized output $\hat{\mathbf{I}}^{lr}_{pix}=D(E(\tilde{\mathbf{I}}^{lr}, \mathbf{M}^{lr}))$ to be close to the downsampled ground-truth real image $\mathbf{I}^{lr}\in \mathbb{R}^{h\times w\times 3}$ by minimizing the reconstruction loss $\mathcal{L}_{pix} = \|\hat{\mathbf{I}}^{lr}_{pix}-\mathbf{I}^{lr}\|_1$.



\subsection{RGB-to-RGB Transformation}\label{rgb}

The simple hand-crafted features used in traditional image harmonization methods have been proven insufficient to acquire appealing harmonization results~\cite{DoveNet2020}. To enable expressive and flexible RGB-to-RGB transformation, we jointly learn a few basis non-linear transformations and a weight predictor to predict the combination coefficient for each transformation. 

For basis transformation, we employ lookup tables (LUTs) due to their simplicity and expressiveness. LUT replaces expensive input/output operation with a simple array indexing operation, and it has been employed to transform input color value to desired output color value in many image processing methods \cite{LUT-inverse-halftoning3,LUT-image-enhancement1,Zeng2020,LUT-image-denoising1,LUT-image-denoising2}. As we expect joint control on the RGB values (3 channels as a whole) instead of a single channel, our used LUT could be regarded as a 3-dimensional grid that defines a conversion matrix in RGB space. As shown in Figure \ref{fig:rgb_trans}, the RGB-to-RGB transformation using LUT could be achieved by two steps. First, given input RGB values $(r,g,b)$, we \emph{look up} its 3D coordinates in LUT. Then \emph{trilinear interpolation} is performed based on its eight nearest surrounding elements to calculate the output RGB values $(r^{\prime},g^{\prime},b^{\prime})$ without sacrificing the continuity of the RGB values. Since the interpolation only introduces a small amount of computation, the combination of these two steps is still computationally efficient. 

Different LUTs could have different output colors, so we employ a set of $N$ learnable LUTs $\{\Phi_n\}_{n=1,..., N}$ as the basis transformations to cover the color transformation space between inconsistent foreground and background regions. Once learned, the basis LUTs are universal for all images. Then, inspired by \cite{Zeng2020}, we employ a weight predictor responsible for image-specific transformation by predicting image-specific combination coefficients. Our weight predictor is built upon the bottleneck feature map in low-resolution harmonization network for the following reasons. Firstly, deep network has a strong capability of capturing the context of input images. Thus the extracted bottleneck feature map contains rich information that is useful for image harmonization. 
Secondly, by sharing the encoder, pixel-to-pixel transformation and RGB-to-RGB transformation are accommodated under a multi-task learning framework, in which two tasks can benefit each other. 

As shown in Figure \ref{fig:network}, the low-resolution generator takes in low-resolution image and mask, and extracts the bottleneck feature map $\mathbf{F}_{enc}=E(\tilde{\mathbf{I}}^{lr}, \mathbf{M}^{lr})$. 
Since the predicted coefficients are expected to adjust the foreground according to the background, we conjecture that explicitly comparing foreground and background features may help learn better coefficients (see Table~\ref{tab:ablate}). 
In particular, based on the mask $\mathbf{M}^{lr}$, we use average pooling to aggregate the $L$-dimensional feature vector $\mathbf{f}_f$ and $\mathbf{f}_b$ for foreground and background separately, which are concatenated as a $2L$-dimensional feature vector $\mathbf{f}$. 
Then, we apply a simple fully connected layer to $\mathbf{f}$ to obtain the image-specific coefficients $\{w_n\}_{n=1,...,N}$, where $N$ is the number of basis LUTs. 
For a high-resolution composite image $\tilde{\mathbf{I}}^{hr}$, given $N$ learnable LUTs $\{\Phi_n\}_{n=1,...,N}$ and combination coefficients, its harmonization output of RGB-to-RGB transformation is obtained as
\begin{eqnarray}\label{equ:RGB-to-RGB transformation}
\hat{\mathbf{I}}^{hr}_{rgb} = (\sum_{n=1}^{N}w_n\Phi_n)(\tilde{\mathbf{I}}^{hr}).
\end{eqnarray}
Note that in RGB-to-RGB transformation, the background also remains the same after harmonization. 
To enforce the high-resolution output to be close to the high-resolution ground-truth real image $\mathbf{I}^{hr}$, we employ the reconstruction loss $\mathcal{L}_{rgb} = \|\hat{\mathbf{I}}^{hr}_{rgb}-\mathbf{I}^{hr}\|_1$.

\subsection{Light-weighted Refinement}\label{refine}

After RGB-to-RGB transformation and pixel-to-pixel transformation, we can obtain RGB-to-RGB result $\mathbf{\hat I}^{hr}_{rgb}\in \mathbb{R}^{H\times W\times 3}$ and pixel-to-pixel result $ \mathbf{\hat I}^{lr}_{pix}\in \mathbb{R}^{h\times w\times 3}$. $ \mathbf{\hat I}^{hr}_{rgb}$ is with high-resolution but insufficient local context. $\mathbf{\hat I}^{lr}_{pix}$ is with low-resolution but rich local context. Hence, they are complementary with each other. However, naively mixing $\mathbf{\hat I}^{hr}_{rgb}$ and $\mathbf{\hat I}^{lr}_{pix}$ can only lead to a blurry and unsatisfactory output, so we design a light-weighted refinement module $R$ to generate better high-resolution result $\mathbf{\hat I}^{hr} \in \mathbb{R}^{H\times W\times 3}$. 

Specifically, we first concatenate bilinearly upsampled $Up(\hat{\mathbf{I}}^{lr}_{pix})\in \mathbb{R}^{H\times W\times 3}$ and $\hat{\mathbf{I}}^{hr}_{rgb}\in \mathbb{R}^{H\times W\times 3}$ in channels. As stated in  \cite{bargain}, mask is essential for image harmonization by explicitly indicating the foreground region. Additionally, the last decoder feature map $\mathbf{F}_{dec}$ accounts for both harmonized foreground and soft attention mask, so we conjecture that $\mathbf{F}_{dec}$ contains rich prior knowledge of both harmonization and local context. 
Therefore, we further append the binary mask $\mathbf{M}^{hr}\in \mathbb{R}^{H\times W\times 1}$ and bilinearly upsampled $Up(\mathbf{F}_{dec})\in \mathbb{R}^{H\times W\times c}$ to the input.

Our refinement module $R$ contains two convolution layers with kernel 3 and stride 1, each followed by a batch normalization and an ELU. At the end of $R$, we also employ an image blending layer to blend both the high-resolution harmonized foreground and input composite image $\tilde{\mathbf{I}}^{hr}$.
Consequently, the refinement module $R$ receives the $H\times W\times (c+7)$ input and produces the better refined high-resolution output $\hat{\mathbf{I}}^{hr} \in \mathbb{R}^{H\times W\times 3}$. We impose the reconstruction loss to enforce the $\hat{\mathbf{I}}^{hr}$ to be close to the high-resolution ground-truth real image $\mathbf{I}^{hr}$, which is denoted by $
\mathcal{L}_{ref} =\|\hat{\mathbf{I}}^{hr}-\mathbf{I}^{hr}\|_1$.

Therefore, the overall loss function of our CDTNet is
\begin{equation}\label{equ:all}
\mathcal{L} = \mathcal{L}_{pix}+\mathcal{L}_{rgb}+\mathcal{L}_{ref}.
\end{equation}

Despite the simple structure of the refinement module, it performs well in fusing the RGB-to-RGB result and the pixel-to-pixel result due to the informative input and the blending layer, which will be validated in Table~\ref{tab:ablate}.

\begin{table*}[h]
\small
\centering
\resizebox{\linewidth}{!}{
    \begin{tabular}{c|c|cccc}
    \hline
    \makecell[c]{Image\\Size} & Method & MSE$\downarrow$  & PSNR$\uparrow$ \ & fMSE$\downarrow$ & SSIM$\uparrow$ \\
    \hline \multirow{11}*{\makecell[c]{1024 \\
    $\times$\\1024}} &  Composite images & 352.05 & 28.10 & 2122.37 & 0.9642 \\
    \cline{2-6} 
    ~ & pix2pixHD \cite{Wang2018} & 63.45 &	31.64 &	332.43 & 0.9135\\
    ~ & CRN \cite{Chen2017} & 90.11	& 29.77	& 259.28	& 0.8225 \\
    ~ & HiDT \cite{anokhin2020high} & 265.32 & 	29.95 & 1501.93 & 0.9628 \\
    ~ & DoveNet \cite{DoveNet2020} & 51.00 & 34.81 & 312.88 & 0.9729\\
    ~ & S$^2$AM \cite{xiaodong2019improving} & 47.01 & 35.68 & 262.39 & 0.9784 \\
    ~ & Guo et al.~\cite{guo2021intrinsic} & 56.34 & 34.69 & 417.33 & 0.9471\\
    ~ & RainNet \cite{ling2021region} &42.56 & 36.61 & 305.17 & 0.9844 \\
    ~ & iS$^2$AM \cite{sofiiuk2021foreground} & 25.03 &	38.29 &	168.85 & 0.9846 \\
    \cline{2-6} 
    ~ & CDTNet-256 (sim) & 31.15 & 37.65 & 195.93 & 0.9841\\
    ~ & CDTNet-256 & \bf21.24 & \bf38.77 & \bf152.13 & \bf0.9868 \\
    \hline
    \end{tabular}
\hspace{1em}
    \begin{tabular}{c|c|cccc}
    \hline
    \makecell[c]{Image\\Size} & Method & MSE$\downarrow$  & PSNR$\uparrow$ \ & fMSE$\downarrow$ & SSIM$\uparrow$ \\
    \hline \multirow{6}*{\makecell[c]{2048\\
    $\times$\\2048}} &  Composite images & 353.92 &	28.07 & 2139.97	& 0.9631 \\
    \cline{2-6} 
    ~ & iS$^2$AM \cite{sofiiuk2021foreground}  & 46.37 &	36.57 &	271.59 & 0.9838 \\
    \cline{2-6} 
    ~ & CDTNet-256 (sim) & 41.11 & 37.28 & 234.06 & 0.9819 \\
    ~ & CDTNet-256 & 29.02 & 37.66	& 198.85 & 0.9845 \\
    ~ & CDTNet-512 (sim) & 38.31 & 37.05 & 233.44 & 0.9828 \\
    ~ & CDTNet-512 & \bf23.35 & \bf38.45 & \bf159.13 & \bf0.9853 \\
    \hline
  \end{tabular}
}
\caption{Quantitative harmonization performance evaluation of different methods. ``CDTNet-256/512'' means the resolution of the low-resolution generator is $256\times 256$/$512\times 512$, and ``(sim)'' represents the simplified variant with deep RGB-to-RGB transformation only.} 
\label{tab:perform}
\end{table*}

\begin{table*}[tb]
\small
\centering
\begin{tabular}{c|c|c|c|c|c|c|c|c|c|c|c}
\hline
\multirow{2}{*}{Image Size} & \multirow{2}{*}{Method} & \multicolumn{2}{c|}{HCOCO} & \multicolumn{2}{c|}{HAdobe5k} & \multicolumn{2}{c|}{HFlickr} & \multicolumn{2}{c|}{Hday2night} & \multicolumn{2}{c}{All} \\  
\cline{3-12}
~ & ~ & MSE$\downarrow$ & PSNR$\uparrow$ & MSE$\downarrow$ & PSNR$\uparrow$&	MSE$\downarrow$ & PSNR$\uparrow$ & MSE$\downarrow$ & PSNR$\uparrow$ & MSE$\downarrow$ & PSNR$\uparrow$ \\ \hline
\multirow{2}*{\makecell[c]{256 $\times$ 256}} & iS$^2$AM \cite{sofiiuk2021foreground} & 16.48 & 39.16 & 22.60&	37.24&	69.67&	33.56&	40.59&	37.72&	24.65	&37.95 \\ 
\cline{2-12}
~ & CDTNet-256 & 16.25 & 39.15 & 20.62&	38.24&	68.61&	33.55&	36.72&	37.95&	23.75	&38.23 \\ \hline
\end{tabular}
\caption{Quantitative comparison on low-resolution ($256\times 256$) image harmonization on iHarmony4 dataset. The resolution of our low-resolution generator is $256\times 256$. The performance of iS$^2$AM is tested using the publicly released model from \cite{sofiiuk2021foreground}.}
\label{tab:perform_256}
\end{table*}

\begin{table*}[htb]
\small
\centering
\begin{tabular}{c|c|cc|cc|cc}
\hline
Image Size & Method & Time$\downarrow$ (ms) & Reduction & Memory$\downarrow$ (MB) & Reduction & FLOPs$\downarrow$ (G) & Reduction \\
\hline 
\multirow{3}*{\makecell[c]{1024 $\times$ 1024}} &  iS$^2$AM & 14.7 & - & 5148 & - & 239.36 & - \\
~ & CDTNet-256 (sim) & 3.5 & 76.20\% & 144 & 97.20\% & 3.49 & 98.54\%  \\
~ & CDTNet-256 & 10.8 & 26.53\% & 1923 & 62.65\% & 78.05 & 67.40\% \\
\hline
\hline 
\multirow{5}*{\makecell[c]{2048 $\times$ 2048}} & iS$^2$AM  & 31.1 & - & 20592 & - & 957.43 & -  \\
~ & CDTNet-256 (sim) & 3.5 & 88.74\% & 288 & 98.60\% & 3.49 & 99.64\% \\
~ & CDTNet-256 & 10.9 & 64.95\% & 6723 & 67.35\% & 267.10 & 72.10\% \\
~ & CDTNet-512 (sim) & 3.7 & 88.10\% & 574 & 97.21\% & 13.95 & 98.54\%  \\
~ & CDTNet-512 & 11.5 & 63.02\% & 7691 & 62.65\% & 312.19 & 67.40\% \\
\hline
\end{tabular}
\caption{Quantitative efficiency comparison between iS$^2$AM \cite{sofiiuk2021foreground} and our CDTNet.}
\label{tab:effciency}
\end{table*}

\begin{figure*}[htb!]
\begin{center}
\includegraphics[width=1.0\linewidth]{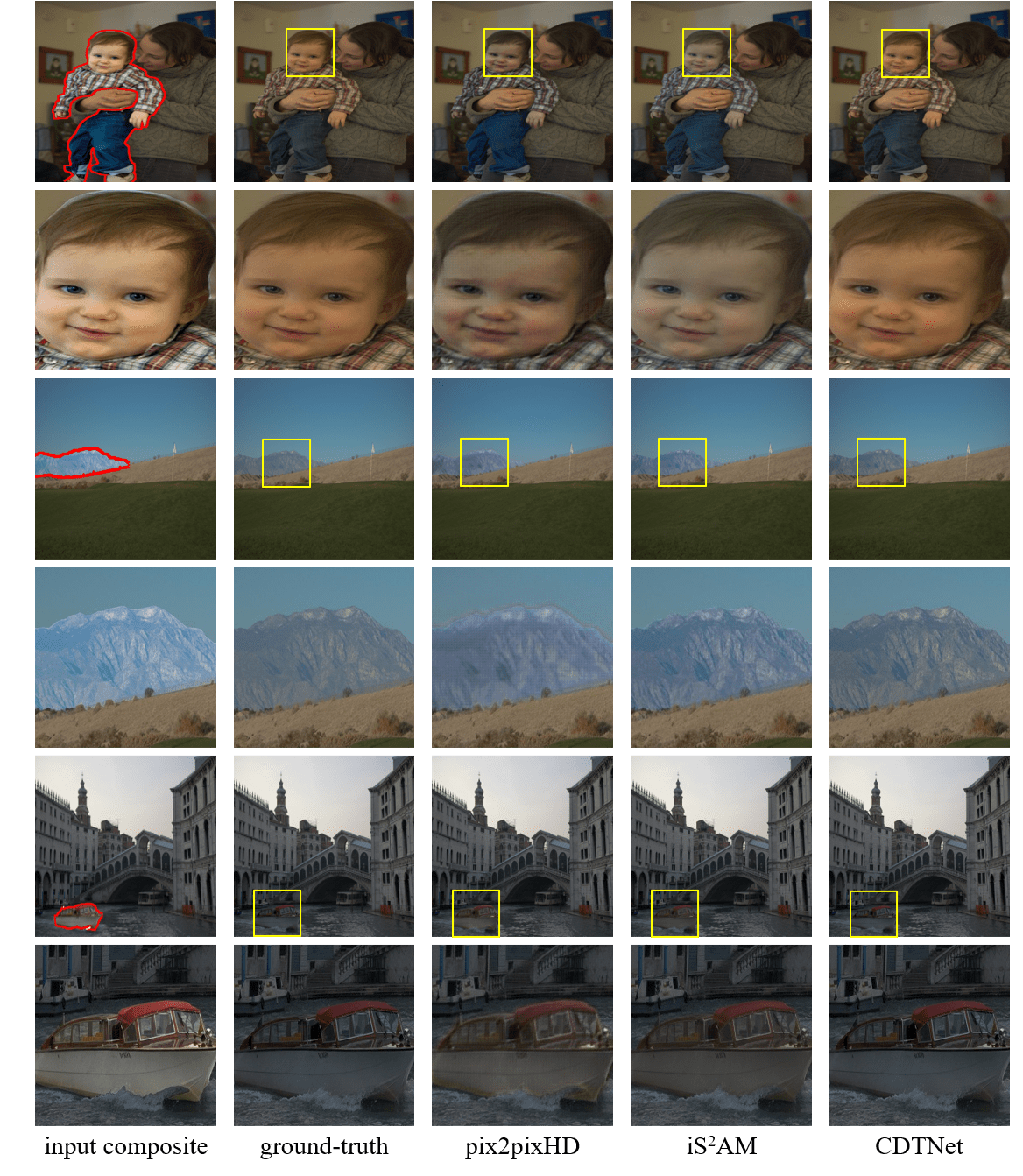}
\end{center}
   \caption{The odd rows show the input composite image, the ground-truth image, as well as example results generated by pix2pixHD~\cite{Wang2018}, iS$^2$AM~\cite{sofiiuk2021foreground}, and our CDTNet on $1024\times1024$ resolution. The red border lines indicate the foreground, and the yellow boxes zoom in the particular regions for a better observation.}
\label{fig:examples}
\end{figure*}

\section{Experiments}\label{exp}

\subsection{Dataset Statistics}\label{data}

Existing image harmonization benchmark iHarmony4 \cite{DoveNet2020} is not specially constructed for high-resolution image harmonization. Among four sub-datasets, HCOCO, HFlickr, and Hday2night are all at a resolution lower than $1024$, and only HAdobe5k contains high-resolution images. Therefore, we conduct experiments on HAdobe5k, which contains 19437 training and 2160 testing pairs of high-resolution composite images and real images.

Considering that the composite images in HAdobe5k are synthesized composite images, we follow \cite{tsai2017deep,DoveNet2020} and further evaluate our model on real composite images. However, existing $99$ real composite images \cite{tsai2017deep,xue2012understanding} are also at a low resolution. Thus, we collect images from Open Image Dataset V6 \cite{oidv6} and Flickr \footnote{https://www.flickr.com}, and create $100$ high-resolution real composite images with diverse foregrounds and backgrounds (see the Supplementary).


\subsection{Implementation Details}\label{implement}

As mentioned in Section~\ref{pix}, we adopt iS$^2$AM backbone proposed in \cite{sofiiuk2021foreground} as the low-resolution generator. In the color mapping module, $\mathbf{f}$ is a $512$-dimensional feature vector with $L=256$. The basis transformations contain $N=4$ learnable LUTs. We also investigate the impact of using different $N$ in Supplementary. 
To ensure that color transformation is well-behaved, we clip the transformed RGB value into the range $[0,1]$. In the refinement module, the number of input channels is $39$ with $c=32$. 
Our network is implemented using Pytorch 1.6.0 and trained using Adam optimizer with learning rate of $1e{-4}$ on ubuntu 18.04 LTS operation system, with 64GB memory, Intel Core i7-8700K CPU, and two GeForce GTX 1080 Ti GPUs. 
\textbf{When conducting experiments on $\mathbf{1024\times1024}$ (\emph{resp.}, $\mathbf{2048\times2048}$) resolution, the resolution of low-resolution generator in our CDTNet is set as 256 (\emph{resp.}, 512) by default.}

We employ four metrics for harmonization performance evaluation, including MSE, foreground MSE (fMSE), PSNR, and SSIM, as well as three metrics for efficiency evaluation, including the average inference time per image, memory cost, and FLOPs. The average inference time per image is evaluated on a single NVIDIA GTX 1080 Ti GPU, and the memory cost and FLOPs are estimated with the network analyzer ``torchstat''.
\newcommand{\tabincell}[2]{\begin{tabular}{@{}#1@{}}#2\end{tabular}}

\subsection{Comparison with Existing Methods}\label{quanti}

Since there are no existing methods specifically designed for high-resolution image harmonization, we compare two baseline groups. The first group contains five recent low-resolution image harmonization methods \cite{DoveNet2020,xiaodong2019improving,sofiiuk2021foreground,guo2021intrinsic,ling2021region}. 
The second group contains high-resolution image-to-image translation methods. We select three representative methods pix2pixHD \cite{Wang2018}, HiDT \cite{anokhin2020high}, and CRN \cite{Chen2017} for comparison. For fairness, we transplant all baselines to high-resolution image harmonization with the slightest modification of their officially released code (see the Supplementary). 
In this section, we refer to our CDTNet with the resolution of low-resolution generator being $256$ (\emph{resp.}, $512$)  as CDTNet-256 (\emph{resp.}, CDTNet-512). 
Moreover, we build a simplified variant (sim) of our CDTNet, which has the same training procedure but only uses deep RGB-to-RGB transformation during inference.  

First, we evaluate the harmonization performance of different methods on $1024\times1024$ resolution.
In Table \ref{tab:perform}, the performances of HiDT and CRN are poor, probably because merging multiple shifted low-resolution results may disturb the pixel values, and generating from repeated refinement may amplify the artifacts. Since high-resolution image-to-image translation baselines are not well-designed for image harmonization task, their overall performance is far from satisfactory. Among the image harmonization baselines, iS$^2$AM achieves competitive performance, which coincides with its superiority at low resolution as reported in \cite{sofiiuk2021foreground} and reveals the reason for employing iS$^2$AM as the low-resolution generator in our method. However, the high-resolution network for pixel-to-pixel transformation may not be good at capturing long-range dependency due to local convolution operations \cite{Wang_2018_CVPR}, especially for the images with large foregrounds (\emph{e.g.}, row 1 in Figure \ref{fig:examples}). The detailed results on different foreground ratio ranges can be found in the Supplementary.
Our CDTNet-256 outperforms \cite{DoveNet2020,xiaodong2019improving,guo2021intrinsic,ling2021region} by a large margin and also beats iS$^2$AM. Even our simplified variant CDTNet-256(sim) outperforms most methods, which demonstrates the expressiveness of our proposed deep RGB-to-RGB transformation.

With the observation that iS$^2$AM~\cite{sofiiuk2021foreground} is the most competitive baseline, we further compare with iS$^2$AM on $2048\times2048$ resolution in Table \ref{tab:perform}. 
The advantage of our method is more obvious, and CDTNet-512 significantly outperforms iS$^2$AM. 
When the resolution of the low-resolution generator is reduced to $256$, our CDTNet-256 still achieves competitive performance and exceeds iS$^2$AM by a large margin on $2048\times2048$ resolution.
For the simplified variants, CDTNet-256 (sim) and CDTNet-512 (sim) both exceed iS$^2$AM, making our proposed deep RGB-to-RGB transformation a strong competitor to high-resolution pixel-to-pixel transformation.

Additionally, to evaluate the performance on low-resolution image harmonization, we also compare with the strongest baseline iS$^2$AM on $256\times 256$ resolution images by using the training set and test set from iHarmony4 dataset \cite{DoveNet2020}. The results in Table \ref{tab:perform_256} show that our CDTNet is slightly better than iS$^2$AM.




In Table \ref{tab:effciency}, we report the inference time, memory cost, and FLOPs when harmonizing a single image in the testing stage. 
Compared to iS$^2$AM, our CDTNet-256 (\emph{resp.}, CDTNet-512) is $26.53\%$ (\emph{resp.}, $63.02\%$) faster with $62.65\%$ less memory cost and $67.40\%$ fewer FLOPs for $1024\times1024$  (\emph{resp.}, $2048\times2048$) images. When the resolution of low-resolution generator is reduced to $256$ on $2048\times2048$, CDTNet-256 requires even less computational resources and costs less time.
For our simplified variants, the superiority of efficiency is more striking, saving $>95\%$ memory and FLOPs and reducing $>75\%$ time, which also demonstrates that RGB-to-RGB transformation is not constrained by the number of pixels and could achieve higher efficiency and lower memory consumption. 



\begin{table}[t]
\small
\centering
\resizebox{\linewidth}{!}{
\begin{tabular}{cccc|cc}
\hline 
Row & Pixel Trans & RGB Trans & Refinement & PSNR$\uparrow$ \ & fMSE$\downarrow$ \\
\hline 
1 & \checkmark & & & 29.41 & 265.2 \\
2 & & \checkmark & & 37.65 & 195.93 \\
3 & & w/o shared $E$ &  & 36.86 & 248.82  \\
4 & & w/o $\mathbf{f}_f\circ \mathbf{f}_b$  & & 37.31 & 217.91  \\
5 & \checkmark &\checkmark& w/o $\hat{\mathbf{I}}^{hr}_{rgb}$ & 37.29 & 226.17  \\
6 & \checkmark&\checkmark& w/o $\hat{\mathbf{I}}^{lr}_{pix}$ &   37.78 & 188.97  \\
7 & \checkmark& \checkmark& w/o $\mathbf{M}^{hr}$ & 37.33 & 202.14   \\
8 & \checkmark & \checkmark & w/o $\mathbf{F}_{dec}$ & 37.64 & 193.90  \\
9 & \checkmark & \checkmark & w/o $B_r$ & 28.13 & 688.55  \\
10 & \checkmark & \checkmark & \checkmark & \bf38.77 & \bf152.13  \\ \hline 
\end{tabular}}
\caption{Ablation studies of low-resolution generator, color mapping module, and refinement module on the $1024\times1024$ resolution. ``$\circ$'' stands for the concatenation, and $B_r$ stands for the blending layer in the refinement module.}
\label{tab:ablate}
\end{table}

\subsection{Ablation Studies}\label{ablation}


Recall that our CDTNet consists of a low-resolution generator for pixel-to-pixel transformation, a color mapping module for RGB-to-RGB transformation, and a refinement module to take advantage of both. Therefore, in this section, we demonstrate the role of low-resolution generator, color mapping module, and refinement module by ablating each component and analyzing different variants with different input types. By taking the $1024\times1024$ resolution as an example, we report the results in Table \ref{tab:ablate}. We can observe that when we only use a low-resolution generator and naively upsample the low-resolution output, it will lead to unsatisfactory and blurry outputs (row 1). 
When only using color mapping module, the result (row 2) is much better than upsampled low-resolution output (row 1), which demonstrates the effectiveness of deep RGB-to-RGB transformation. 
Furthermore, we compare with two variants of color mapping module. Recall that our weight predictor is based on the encoder feature map in low-resolution generator. The first variant is using a separate encoder to extract features for the weight predictor. After using a separate encoder, the performance is downgraded by a large margin (row 3 \emph{v.s.} row 2), demonstrating the effectiveness of jointly performing low-resolution image harmonization and learning RGB-to-RGB transformation. The second variant is using globally pooled encoder feature instead of concatenating the pooled foreground and background features. The performance also drops (row 4 \emph{v.s.} row 2), which shows the advantage of dealing with foreground and background separately.

Then, we ablate each type of input for the refinement module. We ablate the RGB-to-RGB result $\hat{\mathbf{I}}^{hr}_{rgb}$, the upsampled pixel-to-pixel result $\hat{\mathbf{I}}^{lr}_{pix}$, mask $\mathbf{M}^{hr}$, and upsampled low-resolution feature $\mathbf{F}_{dec}$, separately (row 5 to row 8). We can see that the results after removing each type of input all become worse, which indicates the necessity of using all types of input (row 10). Besides, we also remove the blending layer in the refinement module. The obtained result (row 9) is significantly downgraded, which shows that blending layer is very useful for maintaining the background and focusing on adjusting the foreground. 

To better demonstrate the effectiveness of each component, we provide some example images harmonized by different ablated versions in the Supplementary.
Moreover, we investigate the efficiency of each individual module and report the quantitative results (time, memory, FLOPs as in Table \ref{tab:effciency}) in the Supplementary.



\subsection{Qualitative Analyses}\label{quality}

We show the high-resolution ($1024\times 1024$) harmonization results of pix2pixHD \cite{Wang2018}, iS$^2$AM \cite{sofiiuk2021foreground}, and our CDTNet in Figure \ref{fig:examples}. pix2pixHD is not specifically designed for image harmonization, so its performance is less satisfactory. Due to the weak ability to capture long-range dependencies, iS$^2$AM may fail to generate globally harmonious foreground. Compared with them, the results of our model are more plausible and harmonious, which are visually closer to the ground-truth image. More results and analyses are left to the Supplementary.

\subsection{Evaluation on Real Composite Images}\label{sec:real_comp}

As mentioned in Section~\ref{data}, to evaluate the effectiveness of our proposed CDTNet in real scenarios, we manually create $100$ high-resolution real composite images and conduct user study to compare our CDTNet with pix2pixHD \cite{Wang2018} and iS$^2$AM \cite{sofiiuk2021foreground} following \cite{tsai2017deep,DoveNet2020}. The details of user study and harmonization results on real composite images could be found in the Supplementary.




\section{Conclusion}
In this work, we have proposed a novel high-resolution image harmonization method CDTNet with collaborative dual transformations. Our CDTNet consists of a low-resolution generator, a color mapping module, and a refinement module, which integrates pixel-to-pixel transformation and RGB-to-RGB transformation into a unified end-to-end network. 
Extensive experiments on HAdobe5k dataset and real composite images have demonstrated that our method can achieve better harmonization performance with higher efficiency.

\section*{Acknowledgement}
The work is supported by the National Key R\&D Program of China (2018AAA0100704), the National Science Foundation of China (62076162), and the Shanghai Municipal Science and Technology Major Project, China (2021SHZDZX0102).
{\small
\bibliographystyle{ieee_fullname}
\bibliography{egbib}
}

\end{document}


\title{Supplementary Material for High-Resolution Image Harmonization via Collaborative Dual Transformations}

\author{$\textnormal{Wenyan Cong}^{1}$, $\textnormal{Xinhao Tao}^{2}$, $\textnormal{Li Niu}^{1}$\thanks{Corresponding author.}, $\textnormal{Jing Liang}^{1}$, $\textnormal{Xuesong Gao}^{3,4}$, $\textnormal{Qihao Sun}^{4}$, $\textnormal{Liqing Zhang}^{1}$\\
$^1$ Shanghai Jiao Tong University 
$^2$ Harbin Institute of Technology
$^3$ Tianjin University
$^4$ Hisense\\
{\tt\small$^1$\{plcwyam17320,ustcnewly,leungjing\}@sjtu.edu.cn $^2$1180300213@stu.hit.edu.cn}\\
{\tt\small $^{3,4}$gaoxuesong@tju.edu.cn $^4$sunqihao@hisense.com $^1$zhang-lq@cs.sjtu.edu.cn }
}
\maketitle

In this supplementary, we will first introduce how we transplant two baseline groups to high-resolution image harmonization task in Section~\ref{supp_baseline}.
Then, we will investigate the performance on different foreground ratio ranges in Section~\ref{supp_fg_ratio}.
We will conduct ablation studies to analyze the roles of three components in our CDTNet in Section \ref{supp_abalte}, including both qualitative analysis in Section~\ref{supp_perform} and efficiency analysis in Section~\ref{supp_effi}.
Then, we will take an in-depth look at our color mapping module in Section \ref{supp_lut}. Besides, we will provide more visual results of different methods on two different resolutions (\emph{i.e.}, $1024\times1024$ and $2048\times 2048$) in Section \ref{supp_visual}. Furthermore, we will introduce the details of our created $100$ high-resolution real composite images and the conducted user study, and exhibit some harmonized results of different methods on real composite images in Section \ref{supp_realresults}. Finally, we will discuss the limitations of our method in Section~\ref{supp_limit}. \emph{Note that when conducting experiments on $1024\times1024$ (resp., $2048\times2048$) resolution, the resolution of low-resolution generator in our CDTNet is set as 256 (resp., 512), unless otherwise stated. }

\section{Baseline Transplantation}\label{supp_baseline}

Since there are no existing high-resolution image harmonization methods available for comparison, we transplant low-resolution image harmonization methods \cite{DoveNet2020,xiaodong2019improving,sofiiuk2021foreground,guo2021intrinsic,ling2021region} and high-resolution image-to-image translation methods~\cite{Wang2018,anokhin2020high,Chen2017} to our task with essential modification of their official implementation.
The low-resolution image harmonization models \cite{DoveNet2020,xiaodong2019improving,sofiiuk2021foreground,guo2021intrinsic,ling2021region} can be trained on high-resolution images despite the huge memory consumption. Thus, we train these models on high-resolution images with sufficient GPU memory.
For high-resolution image-to-image translation methods, we modify their input by concatenating foreground mask and the composite image, leaving the other components of the network untouched, because the foreground mask has been proved essential for the harmonization task \cite{DoveNet2020,xiaodong2019improving}.

\section{Foreground Ratio Ranges Analyses}\label{supp_fg_ratio}

To better demonstrate that high-resolution pixel-to-pixel transformation may be weak in capturing long-range dependency due to local convolution operations \cite{Wang_2018_CVPR}, we investigate the performance of iS$^2$AM and our CDTNet in different foreground ratio ranges based on MSE and foreground MSE (fMSE) metrics. The results are reported in Table\ref{tab:ablate_ratio}. 
On $1024\times1024$ resolution, iS$^2$AM achieves a comparable performance to our CDTNet when the foreground ratios are less than $15\%$. However, when the foreground ratios are greater than $15\%$, our CDTNet outperforms iS$^2$AM by a large margin.
On $2048\times2048$ resolution,  our CDTNet outperforms iS$^2$AM in all ranges of foreground ratios. Moreover, the performance gap increases as the foreground ratio range increases, which strongly demonstrates that iS$^2$AM tends to have inferior performance especially when the foregrounds are large.


\begin{table*}[tb]
\centering
\begin{tabular}{c|c|cc|cc|cc|cc}
\hline
\multirow{2}*{\makecell[c]{Image Size}} &
\multirow{2}*{\makecell[c]{Foreground ratios}} & \multicolumn{2}{c|}{$0\%\sim 5\%$} & \multicolumn{2}{c|}{$5\%\sim 15\%$} & \multicolumn{2}{c|}{$15\%\sim 100\%$}  & \multicolumn{2}{c}{$0\%\sim 100\%$}\\ \cline{3-10}
~ & ~  & MSE$\downarrow$  & fMSE$\downarrow$   & MSE$\downarrow$  & fMSE$\downarrow$ & MSE$\downarrow$  & fMSE$\downarrow$ & MSE$\downarrow$  & fMSE$\downarrow$   \\ \hline \hline
\multirow{3}*{\makecell[c]{1024
$\times$1024}} & Input composite  & 47.70 & 1810.54 & 179.51 & 1922.07 & 827.27 & 2630.86 & 352.05 & 2122.37  \\ 
~ & iS$^2$AM~\cite{sofiiuk2021foreground} & 4.91 & 192.56 & 13.98 & 151.67 & 56.19 & 162.23 & 25.03 & 168.85 \\
~ & CDTNet  & \bf4.34 & \bf178.50 & \bf13.32 & \bf146.95 & \bf44.80 & \bf132.20 & \bf21.24 & \bf152.13 \\ \hline \hline
\multirow{3}*{\makecell[c]{2048
$\times$2048}} & Input composite  & 48.28 & 1834.87 & 180.91 & 1938.20 & 830.90 & 2642.80 & 353.92 & 2139.97  \\ 
~ & iS$^2$AM~\cite{sofiiuk2021foreground} & 6.51 & 261.54 & 23.47 & 247.15 & 108.99 & 305.52 & 46.37 & 271.59 \\
~ & CDTNet  & \bf3.98 & \bf161.13 & \bf12.67 & \bf140.14 & \bf51.14 & \bf145.64 & \bf23.35 & \bf159.13 \\ \hline
\end{tabular}
\caption{MSE and foreground MSE (fMSE) of iS$^2$AM and our CDTNet in each foreground ratio range based on the whole test set. The best results are denoted in boldface.}
\label{tab:ablate_ratio}
\end{table*}

\section{Ablation Studies}\label{supp_abalte}

\subsection{Qualitative Analyses}\label{supp_perform}

\begin{figure*}[!htp]
\begin{center}
\includegraphics[width=0.94\linewidth]{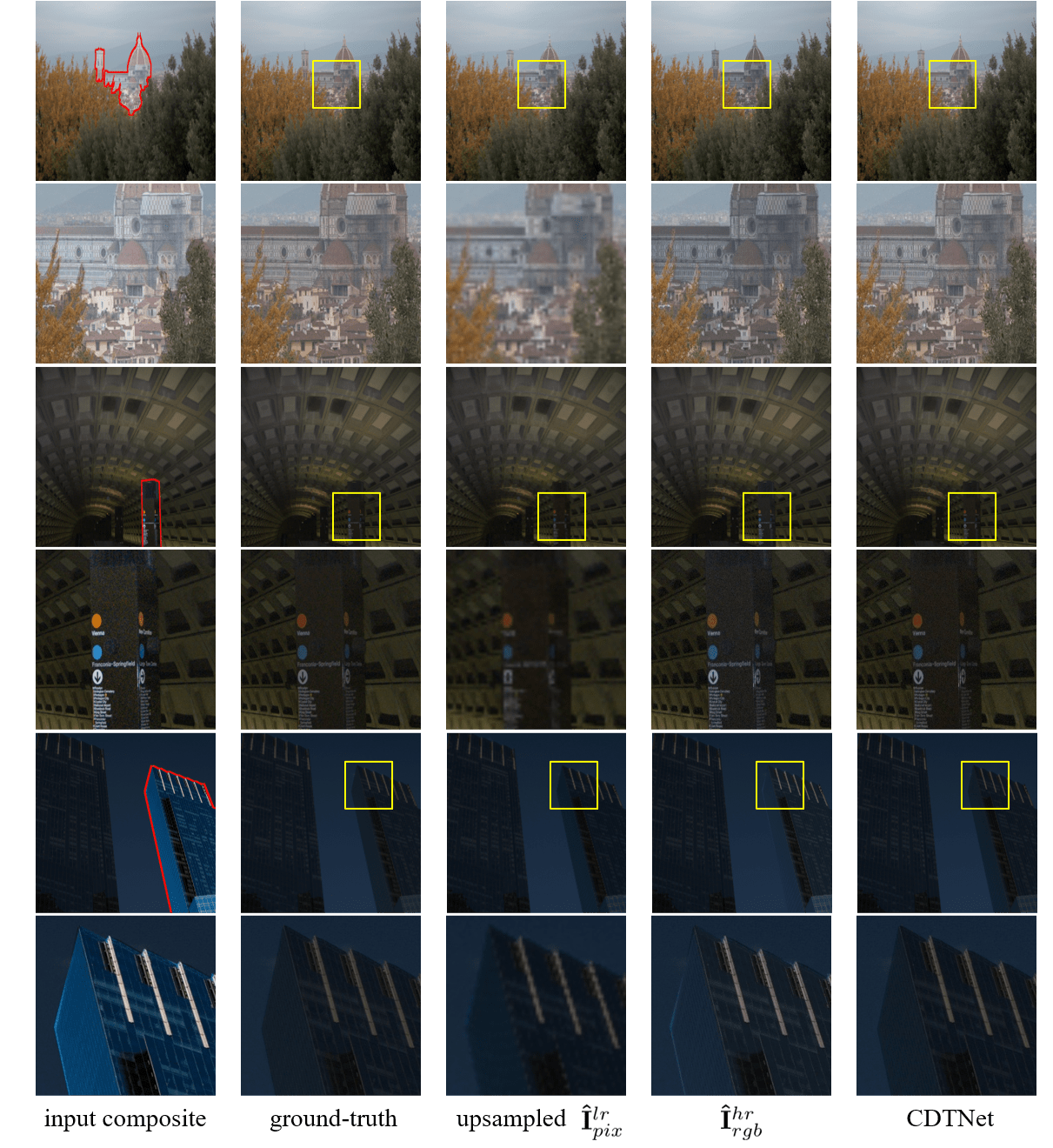}
\end{center}
   \caption{Example results harmonized by only using the low-resolution generator $\hat{\mathbf{I}}^{lr}_{pix}$ (row 1 in Table 4 in the main paper), only using the color mapping module $\hat{\mathbf{I}}^{hr}_{rgb}$ (row 2), and our full method CDTNet (row 10). The red border lines indicate the foreground, and the yellow boxes zoom in the particular regions for a better observation.}
\label{fig:abalate_fig}
\end{figure*}

Our CDTNet consists of a low-resolution generator for pixel-to-pixel transformation, a color mapping module for RGB-to-RGB transformation, and a refinement module to take advantage of both. We have provided the quantitative results of ablating each component in Table 4 in the main paper. 
In Figure \ref{fig:abalate_fig}, we present some example images harmonized by different ablated versions on $1024\times1024$ resolution, including only using the low-resolution generator (row 1 in Table 4 in the main paper), only using the color mapping module (row 2 in Table 4 in the main paper), and our full method (row 10 in Table 4 in the main paper). We can observe that the upsampled results of the low-resolution generator are too blurry to meet the satisfaction of high-resolution image harmonization. In contrast, the results of the color mapping module are with high resolution and sharp contour. However, due to the lack of fine-grained information, global RGB-to-RGB transformation cannot leverage local context and may produce unsatisfactory local results (\emph{e.g.}, darker roofs in row 2, brighter signs and words in row 4, more obvious reflective light in row 6). Our full method takes advantage of both modules and produces more favorable results, which are visually pleasant and closer to the ground-truth.


\begin{figure*}[t]
\begin{center}
\includegraphics[width=0.8\linewidth]{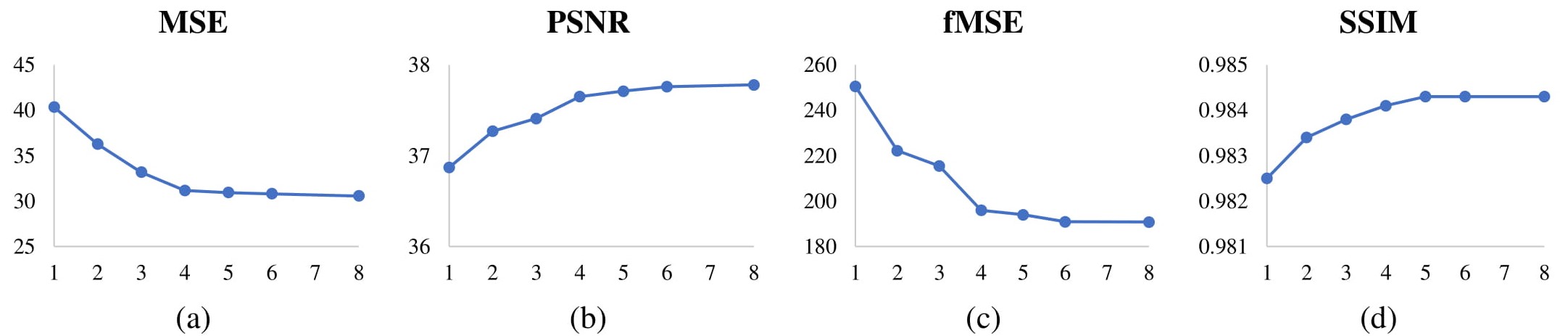}
\end{center}
   \caption{Impact of the number $N$ of the basis transformations on $1024\times1024$ resolution.}
\label{fig:lut_num}
\end{figure*}

\subsection{Efficiency Analyses}\label{supp_effi}

In Table 3 in the main paper, we evaluate the efficiency of our method and its simplified variant. 
Here, we further investigate the efficiency for each individual module (\emph{i.e.}, low-resolution generator \textbf{G}, color mapping module \textbf{C}, and refinement module \textbf{R}) in Table \ref{tab:moduletime}. 

Since the generator only operates on low-resolution inputs, its time consumption, as well as computational and memory cost, are well-surpressed compared to that of iS$^2$AM~\cite{sofiiuk2021foreground} in Table 3 in the main text. Note that the color mapping module in Table \ref{tab:moduletime} does not include the encoder $E$ from the low-resolution generator.
Since global RGB-to-RGB transformation is barely constrained by the number of pixels and the weight predictor is light-weighted in structure, our color mapping module is very efficient in high resolutions.
Our refinement module is also simple in structure. Although its memory cost and FLOPs are relatively higher, the inference time is fast, and the resolution change only causes a slight increase in the time cost.

\begin{table}[tb]
\centering
\begin{tabular}{c|c|ccc}
\hline Image Size & Module & Time$\downarrow$ & Memory$\downarrow$ & FLOPs$\downarrow$ \\
\hline \multirow{3}*{\makecell[c]{1024
$\times$1024}} &  \textbf{G}  & 8.6& 322 &  14.96\\
~ & \textbf{C} & 1.1 & 49& 0.07 \\
~ & \textbf{R} & 1.1 & 1552&  63.02\\
\hline
\hline \multirow{3}*{\makecell[c]{2048
$\times$2048}} & \textbf{G} &9.2 & 1287&  59.84\\
~ & \textbf{C} & 1.1&196 &  0.27\\
~ & \textbf{R} & 1.2  &6208 & 252.07 \\
\hline
\end{tabular}
\caption{Efficiency evaluation for each individual module of our CDTNet on $1024\times1024$ and $2048\times2048$ resolutions, including ``Time'' (ms), ``Memory'' (MB), and ``FLOPs'' (G). \textbf{G}: low-resolution generator. \textbf{C}: color mapping module. \textbf{R}: refinement module.}
\label{tab:moduletime}
\end{table}

\begin{figure*}[tp]
\begin{center}
\includegraphics[width=0.9\linewidth]{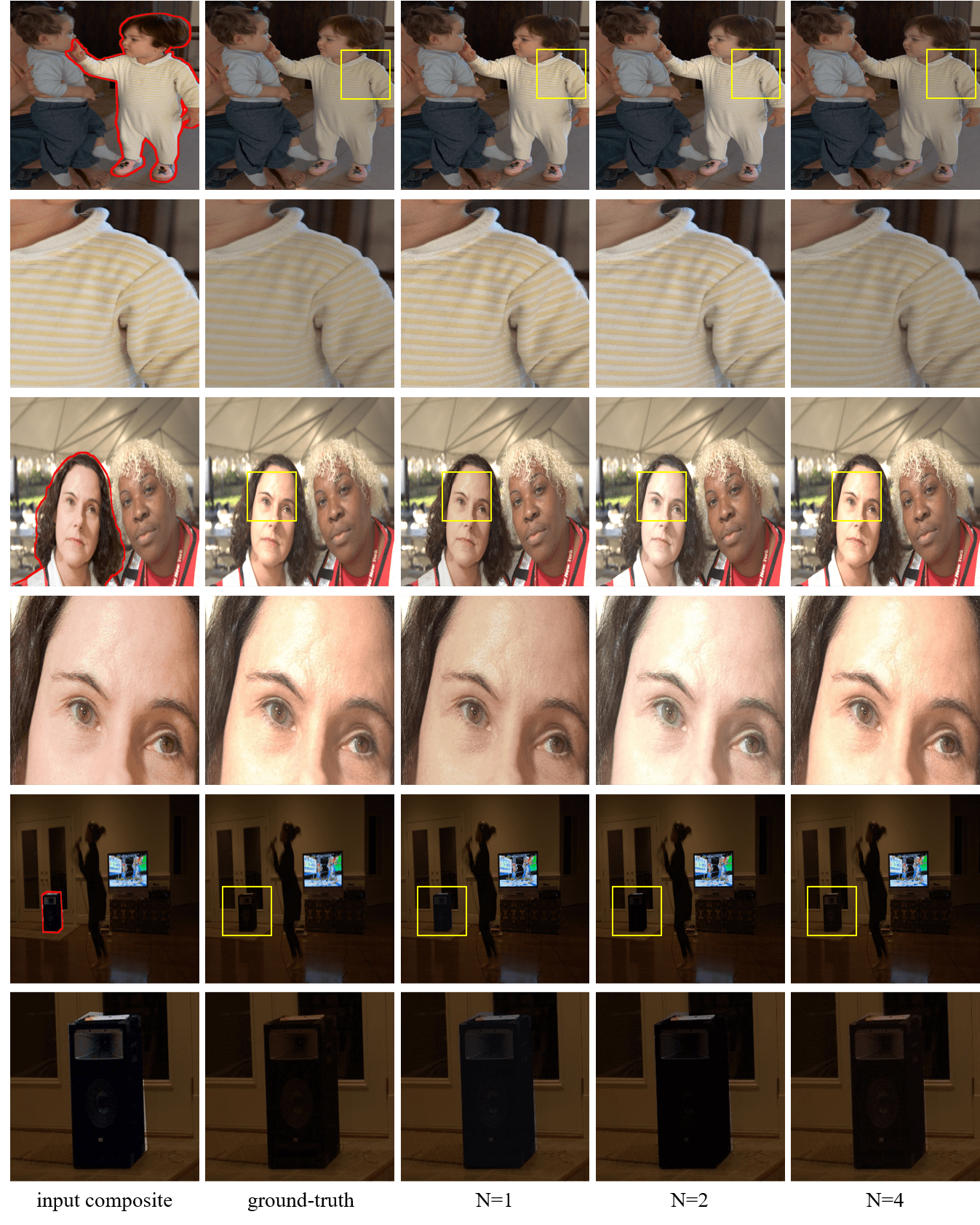}
\end{center}
   \caption{Example harmonized results using different numbers of LUTs in the color mapping module. From left to right, we show the input composite image, the ground-truth image, as well as the harmonized results generated using 1 LUT, 2 LUTs, and 4 LUTs (our setting) on $1024\times1024$ resolution.}
\label{fig:lutvisual}
\end{figure*}

\section{Analyses of Our Color Mapping Module}\label{supp_lut}

In our color mapping module, we employ $N$ basis transformations (LUTs) and a weight predictor to predict the combination coefficients of basis transformations. In the main paper, we set $N=4$ by default. In this section, we first investigate the number $N$ of the basis transformations on $1024\times1024$ resolution. Specifically, we set $N=\{1,2,3,4,5,6,7,8\}$ and employ weight predictor as proposed in Section 3.2 in the main text. 

In Figure \ref{fig:lut_num}, we plot the performance by varying $N$. It can be seen that using only a single LUT has poor performance due to the weak transformation ability. When increasing $N$ from 1 to 4, the performance is boosted obviously because more LUTs can improve the expressiveness of image-specific color transformation. Further increasing $N$ from 4 to 8 leads to the performance convergence with only minor improvements. Since more LUTs will increase the memory consumption, we set $N=4$ by default in all experiments for a good trade-off between performance and memory consumption.

We also present example images harmonized using different numbers $N$ of LUTs, where $N$ is set as $\{1,2,4\}$ and the results are shown in Figure \ref{fig:lutvisual}. It can be seen that as $N$ increases, the harmonized results become more visually appealing and closer to the ground-truth image, because more basis LUTs make the combined transformation more expressive. 

We also take an in-depth look at the learnt $N$ LUTs and observe the transformed result using each LUT. One interesting observation is that when $N>1$, the transformed result using the first LUT is close to the composite image while the transformed results using the other $N\!-\!1$ LUTs look like residues. This might be caused by our way of initializing $N$ LUTs. Specifically, the first LUT is initialized as an identity map while the other $N\!-\!1$ LUTs are initialized as zero maps.  
Therefore, the combination of transformed results using $N$ LUTs is equivalent to making adjustments for the composite image by adding proper residues.


\section{More Visualization Results on HAdobe5k}\label{supp_visual}

\begin{figure*}[tp]
\begin{center}
\includegraphics[width=\linewidth]{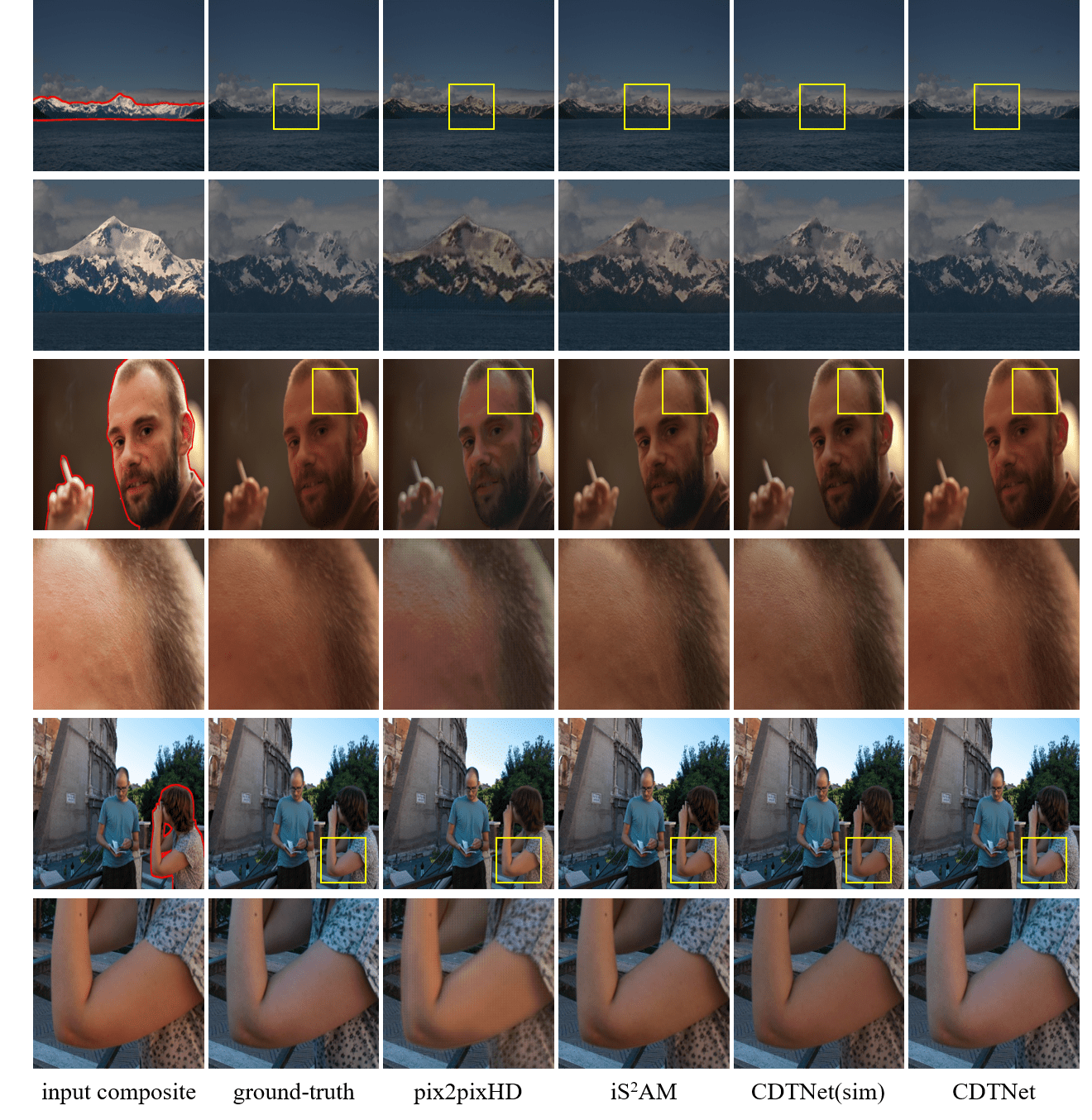}
\end{center}
   \caption{Odd rows show the input composite image, the ground-truth image, as well as example results generated by pix2pixHD~\cite{Wang2018}, iS$^2$AM~\cite{sofiiuk2021foreground}, our simplified variant, CDTNet (sim), which uses deep RGB-to-RGB transformation only, and our CDTNet on $1024\times1024$ resolution. The red border lines indicate the foreground, and the yellow boxes zoom in the particular regions for a better observation.}
\label{fig:example1024}
\end{figure*}

\begin{figure*}[tp]
\begin{center}
\includegraphics[width=0.95\linewidth]{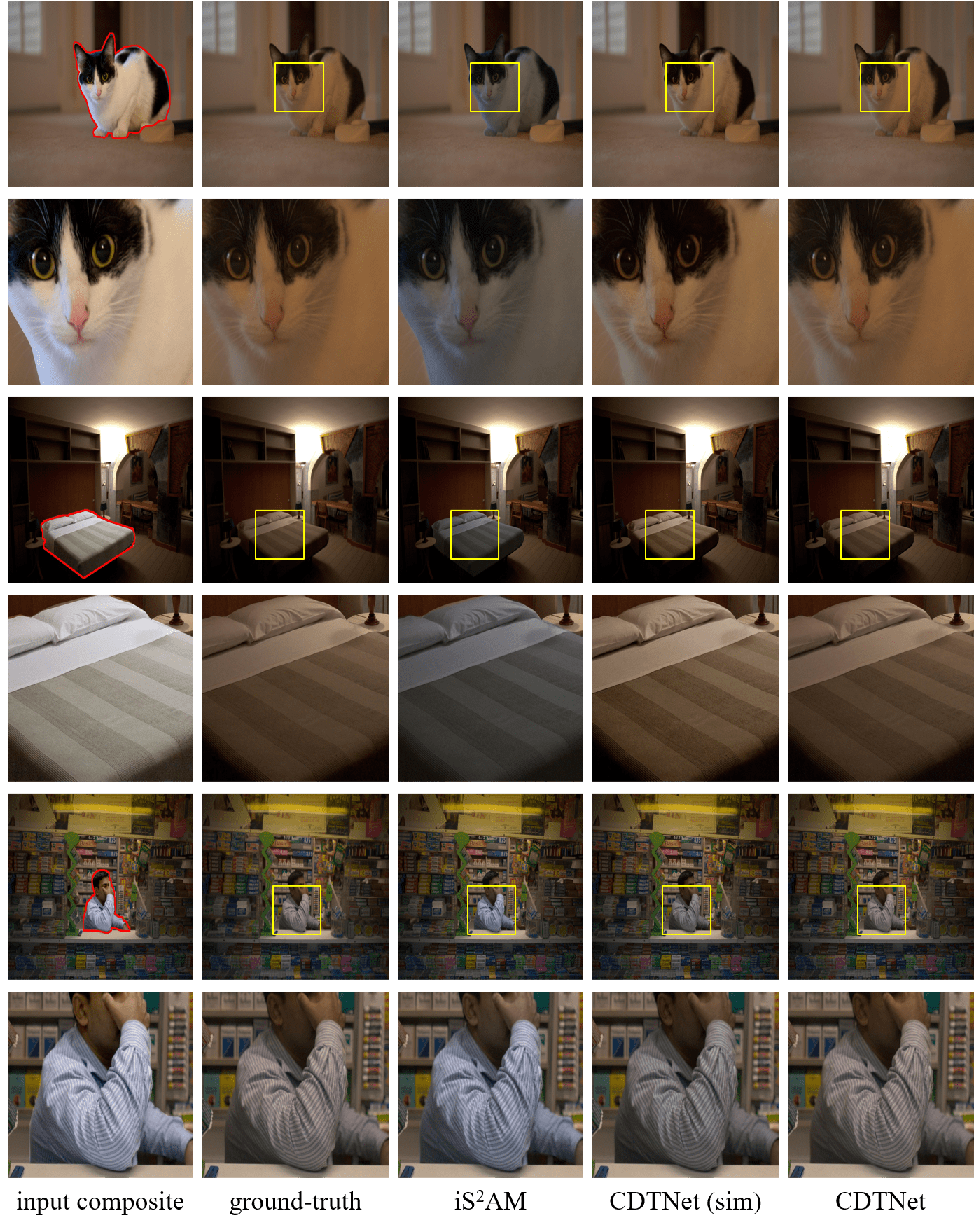}
\end{center}
  \caption{Odd rows show the input composite image, the ground-truth image, as well as example results generated by iS$^2$AM~\cite{sofiiuk2021foreground}, our simplified variant, CDTNet (sim), which uses deep RGB-to-RGB transformation only, and our CDTNet on $2048\times2048$ resolution. The red border lines indicate the foreground, and the yellow boxes zoom in the particular regions for a better observation.}
\label{fig:example2048}
\end{figure*}

We provide more results of baseline pix2pixHD~\cite{Wang2018}, iS$^2$AM~\cite{sofiiuk2021foreground}, our simplified variant which uses deep RGB-to-RGB transformation only (CDTNet-256 (sim) in Table 1 in the main text), and our CDTNet on $1024\times1024$ resolution in Figure \ref{fig:example1024}. We also provide additional results of iS$^2$AM, our simplified variant (CDTNet-512 (sim) in Table 1 in the main text), and our CDTNet on $2048\times2048$ resolution in Figure \ref{fig:example2048}. pix2pixHD~\cite{Wang2018} is not specifically designed for image harmonization, so its performance is less satisfactory. The large image harmonization models directly trained on high-resolution images  may be weak in capturing long-range dependency, as discussed in Section \ref{supp_fg_ratio}.
In Figure \ref{fig:example1024}, our simplified variant (CDTNet (sim)) obtains globally reasonable illumination but insufficient local harmony, while in Figure \ref{fig:example2048}, CDTNet (sim) outperforms iS$^2$AM by generating more harmonious results, which demonstrates the expressiveness of our deep RGB-to-RGB transformation.
Based on Figure \ref{fig:example1024} and Figure \ref{fig:example2048}, for both resolutions, our CDTNet could generate more plausible and satisfactory harmonization results stably and adaptively, which demonstrates the superiority and robustness of our method.

\section{Results on High-Resolution Real Composite Images}\label{supp_realresults}
Considering that the composite images in HAdobe5k are synthesized composite images, we further perform evaluation on $100$ high-resolution real composite images.

\subsection{Image Statistics}\label{supp_hrrealcomp}
We create high-resolution real composite images using the images from Open Image Dataset V6 \cite{OpenImages2}.  
Open Image Dataset V6 contains $\sim$9M images with 28M instance segmentation annotations of 350 categories, where enormous images are collected from Flickr \footnote{https://www.flickr.com} and with high resolution. Therefore, we collect the foreground images from the whole Open Image Dataset V6 and use the provided instance segmentation masks to crop the foregrounds. To ensure the diversity and quality of the composite images, we collect the background images from both Open Image Dataset V6 and Flickr, considering the resolutions and semantics. 
Then, we use PhotoShop to combine cropped foregrounds and background images by placing the foreground region at a reasonable location with a suitable scale. After that, we choose $100$ high-resolution real composite images with obviously inharmonious foreground and background  for evaluation. 

\begin{figure*}[htb]
\begin{center}
\includegraphics[width=0.8\linewidth]{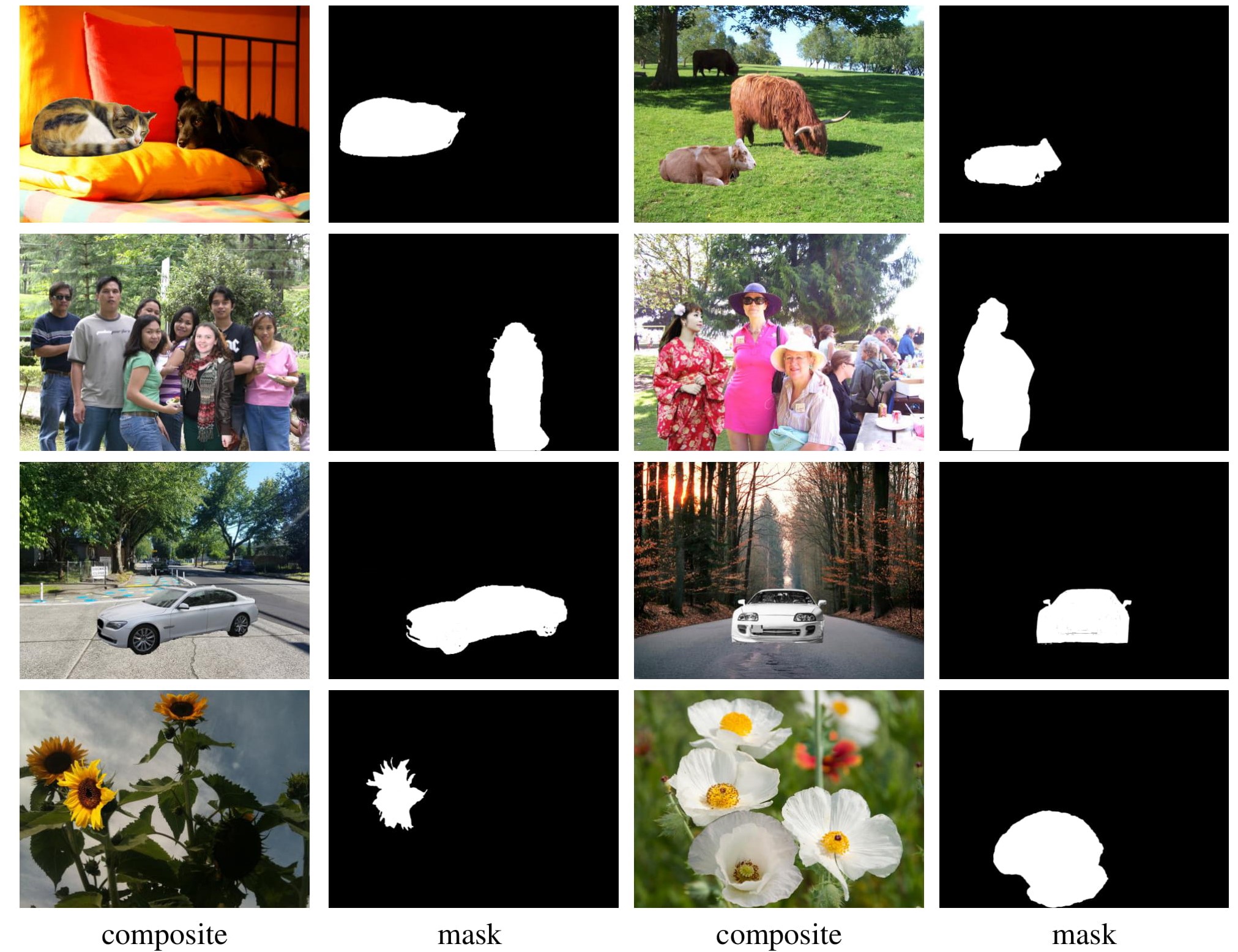}
\end{center}
   \caption{Example composite images and corresponding masks from our created $100$ high-resolution real composite images.}
\label{fig:realcomp}
\end{figure*}

The generated real composite images are with random resolution from 1024 to 6016. The foregrounds include human portraits and general objects (\emph{e.g.}, dog, cat, car), and the backgrounds cover diverse scenes. Example high-resolution real composite images and corresponding masks could be found in Figure \ref{fig:realcomp}.

\begin{table}
\centering
\begin{tabular}{c|c}
\hline Method & B-T score$\uparrow$ \\
\hline
Composite & 0.999 \\
pix2pixHD \cite{Wang2018} & 0.386 \\
iS$^2$AM \cite{sofiiuk2021foreground} & 1.076  \\
CDTNet & 1.216 \\
\hline
\end{tabular}
\caption{B-T scores of baseline pix2pixHD~\cite{Wang2018}, iS$^2$AM~\cite{sofiiuk2021foreground}, and our CDTNet on $100$ high-resolution real composite images. }
\label{tab:BT_score}
\end{table}


\subsection{User Study}\label{supp_userstudy}

To demonstrate the effectiveness of our proposed CDTNet in real scenarios, we follow \cite{tsai2017deep,xiaodong2019improving,DoveNet2020} and further compare our model with baseline pix2pixHD~\cite{Wang2018}, iS$^2$AM~\cite{sofiiuk2021foreground} on $100$ real composite images resized to $1024\times1024$ resolution. More specifically, given each composite image and its 3 harmonized outputs from 3 different methods, we can construct image pairs $(I_i, I_j)$ by randomly selecting two from these 4 images $\{I_i|_{i=1}^4\}$. Hence, we can construct $600$ image pairs based on $100$ real composite images.

Each user involved in this subjective evaluation could see an image pair each time to decide which one looks more harmonious and realistic. Considering the user bias, 14 users participate in the study in total, contributing 8400 pairwise results. With all pairwise results, we employ the Bradley-Terry (B-T) model \cite{bradley1952rank,lai2016comparative} to obtain the global ranking of all methods, and the results are reported in Table \ref{tab:BT_score}. Our proposed method shows an advantage over other methods with the highest B-T score, which demonstrates that by combining the complementary pixel-to-pixel transformation and RGB-to-RGB transformation, our method could generate more favorable results in real-world applications.

\begin{figure}[htb]
\begin{center}
\includegraphics[width=\linewidth]{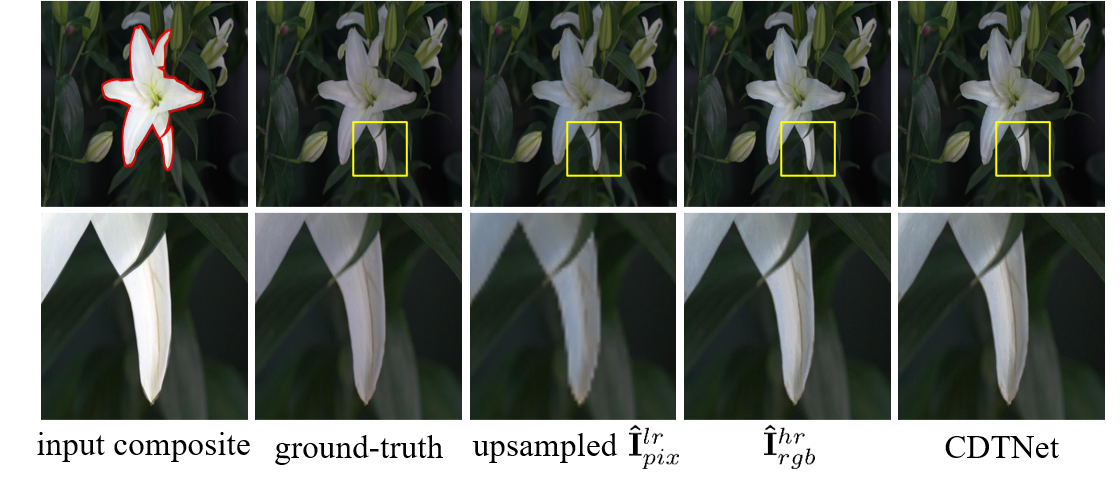}
\end{center}
   \caption{A Sample of failure cases on $1024\times1024$ resolution. From left to right, we show the input composite image, the ground-truth image, as well as the harmonized result generated using only the low-resolution generator, only the color mapping module, and our CDTNet.}
\label{fig:failure}
\end{figure}

\subsection{Qualitative Results}\label{supp_realvisual}

\begin{figure*}[tp]
\vspace{-1em}
\centering
\includegraphics[width=0.85\linewidth]{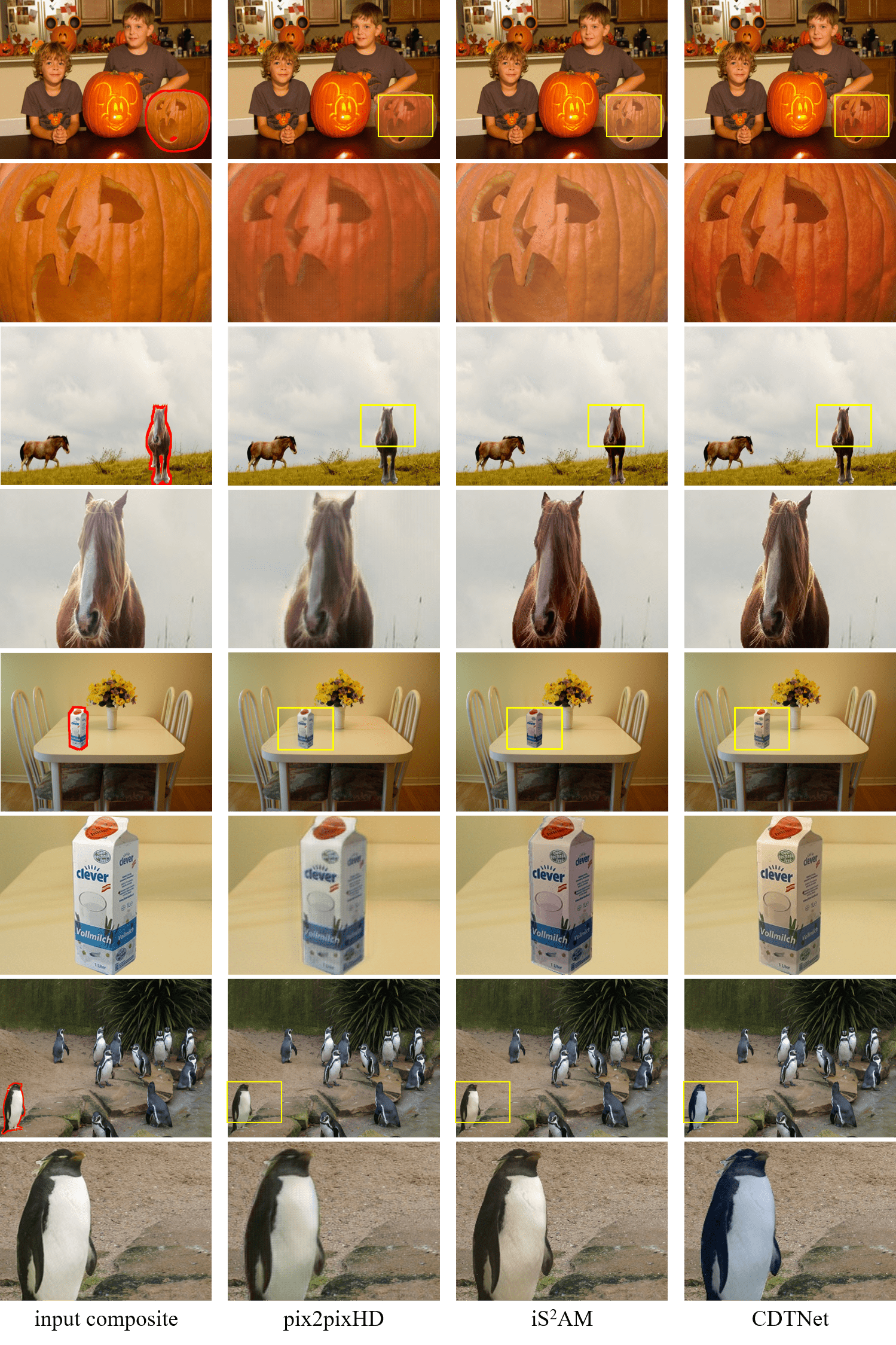}
\vspace{-1em} 
   \caption{Example results on high-resolution real composite images. From left to right, we show the input composite image and the harmonized results generated by pix2pixHD~\cite{Wang2018}, iS$^2$AM~\cite{sofiiuk2021foreground}, and our CDTNet on $1024\times1024$ resolution. The red border lines indicate the foreground, and the yellow boxes zoom in the particular regions for a better observation. All images are resized in this figure for better visualization.}
 
\label{fig:realcompres}
\end{figure*}

\begin{figure*}[tp]
\vspace{-0.5em}
\centering
\includegraphics[width=0.85\linewidth]{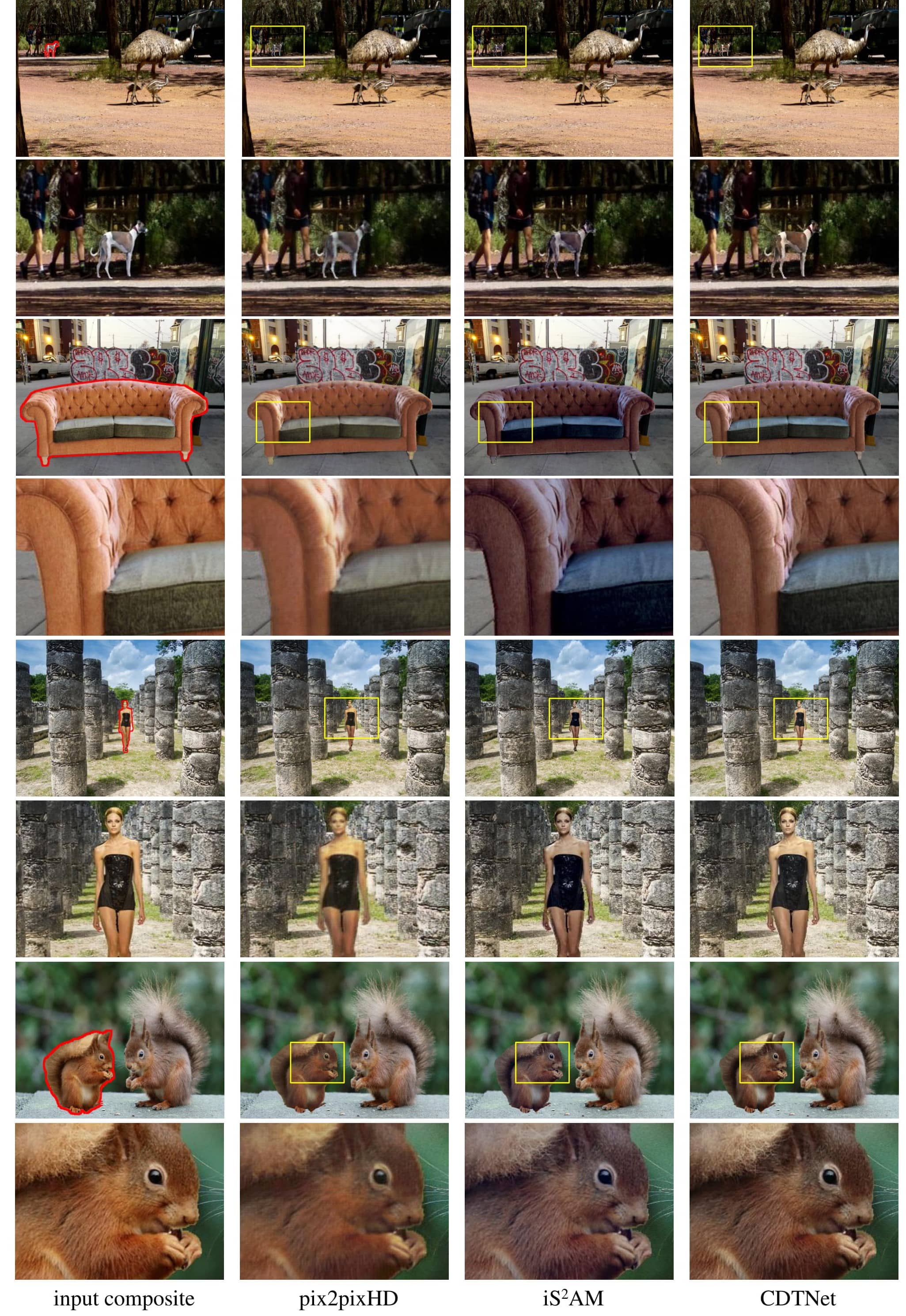}
\vspace{-0.5em}
   \caption{Example results on high-resolution real composite images. From left to right, we show the input composite image and the harmonized results generated by pix2pixHD~\cite{Wang2018}, iS$^2$AM~\cite{sofiiuk2021foreground}, and our CDTNet on $1024\times1024$ resolution. The red border lines indicate the foreground, and the yellow boxes zoom in the particular regions for a better observation. All images are resized in this figure for better visualization.}
\label{fig:realcompres2}
\end{figure*}

To visualize the comparison on high-resolution real composite images, we provide the harmonization results of pix2pixHD~\cite{Wang2018}, iS$^2$AM~\cite{sofiiuk2021foreground}, and our CDTNet in Figure \ref{fig:realcompres} and Figure \ref{fig:realcompres2}. Since pix2pixHD is not well-designed for image harmonization, it tends to produce checkerboard artifacts and halo artifacts in the foreground region, which is especially obvious when zooming in. Therefore, the results of pix2pixHD are far from satisfactory in the real scenario. Compared with iS$^2$AM, our CDTNet is more capable of generating harmonious outputs in real scenarios, which demonstrates the superiority and generalizability of our proposed method.

\section{Limitations}\label{supp_limit}

Although our model could achieve stable and effective image harmonization performance on different resolutions, it might encounter failures with unrealistic local results. For example, in Figure \ref{fig:failure}, the petal in the input composite image is overexposed, which still remain very bright after applying global RGB-to-RGB transformation (see the RGB-to-RGB result $\hat{\mathbf{I}}_{rgb}^{hr}$). The RGB-to-RGB result, as partial input of the refinement module, may adversely affect the refinement module, leading to inharmonious local results.

{\small
\bibliographystyle{ieee_fullname}
\bibliography{egbib}
}